\documentclass[a4paper]{article}
\usepackage{INTERSPEECH2019}

\usepackage{arydshln}
\usepackage{graphicx}
\usepackage{subcaption}
\usepackage{tikz}
\usepackage{xcolor}

\title{Joint Speech Recognition and Speaker Diarization via Sequence Transduction}
\name{Laurent El Shafey, Hagen Soltau, Izhak Shafran}

\address{Google}
\email{shafey@google.com, soltau@google.com, izhak@google.com}

\date{April 2019}

\begin{document}

\maketitle

\begin{abstract}
Speech applications dealing with conversations require not only recognizing the spoken words, but also determining who spoke when. The task of assigning words to speakers is typically addressed by merging the outputs of two separate systems, namely, an automatic speech recognition (ASR) system and a speaker diarization (SD) system. The two systems are trained independently with different objective functions. Often the SD systems operate directly on the acoustics and are not constrained to respect word boundaries and this deficiency is overcome in an {\it ad hoc} manner. Motivated by recent advances in sequence to sequence learning, we propose a novel approach to tackle the two tasks by a joint ASR and SD system using a recurrent neural network transducer. Our approach utilizes both linguistic and acoustic cues to infer speaker roles, as opposed to typical SD systems, which only use acoustic cues. We evaluated the performance of our approach on a large corpus of medical conversations between physicians and patients. Compared to a competitive conventional baseline, our approach improves word-level diarization error rate from 15.8\% to 2.2\%.
\end{abstract}

\section{Introduction}
\label{sec:intro}

\begin{figure}
    \centering
    \begin{subfigure}[b]{0.25\textwidth}
        \includegraphics[width=\textwidth]{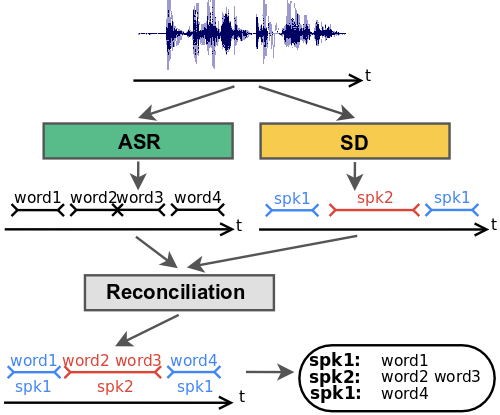}
        \caption{Baseline}
        \label{fig:overview-baseline}
    \end{subfigure}
    \begin{subfigure}[b]{0.2\textwidth}
        \includegraphics[width=\textwidth]{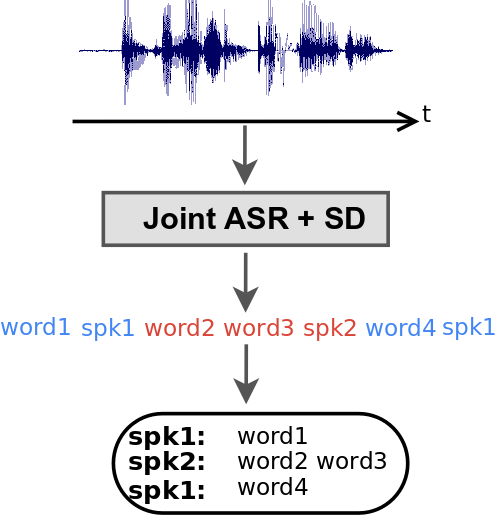}
        \caption{Joint ASR+SD}
        \label{fig:overview-asrd}
    \end{subfigure}
    \caption{Comparison of the conventional speech recognition and speaker diarization system (Figure~\ref{fig:overview-baseline}) with the proposed approach (Figure~\ref{fig:overview-asrd}), where the task consists of generating a speaker-decorated transcript from raw audio.}
    \label{fig:overview}
    \vspace{-0.5cm}
\end{figure}

In the last few decades, speech and language technology has advanced significantly, leading to a profound change in the way people interact with machines and low cost devices. For instance, with the rapid growth of smart speakers, automatic speech recognition (ASR) systems are now commonly used by millions of users. Even with these remarkable advances, machines have  difficulties understanding natural conversations with multiple speakers such as in broadcast interviews, meetings, telephone calls, videos or medical recordings. One of the first steps in understanding natural conversations is to recognize the words spoken and their speakers. As illustrated in Figure~\ref{fig:overview-baseline}, this is typically performed in multiple steps that include (1) transcribing the words using an ASR system, (2) predicting "who spoke when" using a speaker diarization (SD) system, and, finally, (3) reconciling the output of those two systems.

More formally, speaker diarization consists of partitioning an input audio stream into time-bounded segments before annotating each of those segments with a speaker label. Many different SD systems have been proposed in the literature~\cite{Tranter2006,Anguera2012} and they often rely on the following pattern: (a) run a voice activity detector and segment the input audio into speech segments, (b) extract features from each segment to generate a speaker embedding and (c) cluster the resulting speaker embeddings. While early work relied on handcrafted audio features for speaker embedding~\cite{Ajmera2003,Barras2006}, recent efforts have been successful in learning better representations automatically using i-vectors~\cite{Sell2014} in feed-forward neural networks~\cite{Garcia2017} or recurrent neural networks (RNN)~\cite{Wang2018}. The embeddings have been further improved by explicitly maximizing speaker classification accuracy~\cite{Snyder2018,Sell2018}. Triplet loss was proposed as an alternative cost function though the training requirements are cumbersome~\cite{Bredin2017}. In one variant, the clustering step has been successfully replaced with a supervised approach~\cite{Zhang2018}. One commonality with most of the previous work, is that they rely solely on acoustic information to assign speaker labels to audio segments.
% TODO(shafey): extend a little bit this related work paragraph?

However, in many speech applications, subjects in a conversation play very specific roles, which is reflected in their language use. For example, in a clinical visit, the physicians ask all the questions about the symptoms that the patient may be experiencing. Similarly, the patients are likely to be the ones seeking clarifications about treatments. Speaker labels could possibly be inferred from the transcript directly. Only a few approaches have utilized such linguistic cues for speaker diarization. In one approach, linguistic cues were used to associate speaker labels to segments generated by a conventional SD system~\cite{Canseco2004}. More recently, a gated recurrent unit-based sequence to sequence model was employed to detect speaker changes~\cite{Park2018}. The changes were predicted over $32$ word sliding windows and the predictions from overlapping windows were resolved using a voting mechanism, which was then followed by a traditional clustering step.

We propose a novel method to perform automatic speech recognition and speaker diarization in a joint manner, as illustrated in Figure~\ref{fig:overview-asrd}. Our approach utilizes both acoustic and linguistic cues, and is, hence, designed to perform well in scenarios where the speakers involved have well-defined roles. Specifically, our main contribution consists of defining the joint ASR and SD task as a sequence transduction problem and implementing the solution using a recurrent neural network transducer (RNN-T) (Section~\ref{sec:transduction}). We train and evaluate this system on a large corpus of clinical conversations, which is a very good fit for such a joint acoustic and linguistic SD system. The experimental results highlight significant improvements compared to a strong baseline using a conventional SD system (Section~\ref{sec:expts}). Finally, we summarize our findings and provide future lines of research (Section~\ref{sec:conclusions}).

\section{Diarization via Sequence Transduction}
\label{sec:transduction}

\subsection{Problem Formulation and Proposed Solution}
\label{ssec:problem-formulation}

\begin{figure}
\begin{quotation}
  {\color[HTML]{4285F4} hello dr jekyll \texttt{<spk:pt>}} {\color[HTML]{DB4437} hello mr hyde what brings you here today \texttt{<spk:dr>}} {\color[HTML]{4285F4} I am struggling again with my bipolar disorder \texttt{<spk:pt>}}
\end{quotation}
\caption{Example of an output sequence for our joint ASR and SD RNN-T system. The corresponding input would be the raw audio signal. Speaker turns are displayed in different colors.}
\label{fig:output-sequence}
% \vspace{-0.5cm}
\end{figure}

Many machine learning tasks can be expressed as mapping an input sequence into an output sequence. Specifically, speech recognition can be defined as a transformation that outputs a sequence of words from an audio signal. RNNs are popular models that have been used to model such sequential data. In speech recognition, they are often used in a hybrid setting, where alignments are precomputed via the Viterbi algorithm using an existing model~\cite{Robinson1996}. As an alternative, Connectionist temporal classification (CTC) does not rely on pre-computed alignments and maximizes the likelihood via executing the forward-backward algorithm at every step~\cite{Graves2006}. To accommodate different sequence length between input and output, a blank symbol is introduced and the loss is calculated by marginalizing over all possible alignments. The main shortcoming of CTC is that output dependencies are not modeled. RNN-T models can be seen as an extension of CTC that addresses this shortcoming and adds a language model component~\cite{Graves2012}. Initially, the scores from the acoustic and language model components were simply multiplied. This was later improved up on by using a general feed forward layer~\cite{Graves2013} and the resulting model proved to be very successful in various sequence modeling tasks.

Compared to conventional systems where acoustic and language models are trained separately, RNN-T models have recently been successfully applied to speech recognition using end-to-end training, leading to comparable or even higher accuracy on a diverse set of acoustic conditions~\cite{Graves2013,He2018}. 

% Commenting this out to remove redundancy, for page limit
% Motivated by this approach, we propose to extend it one step further by not only outputting words but also speaker tokens at the end of each turn. 
% However, their proposed system doesn't perform the ASR and SD task jointly, but instead is fed by both acoustic features and the corresponding ASR words as inputs, and outputs an interleaved sequence of words and speaker labels. 
% In contrast, we propose a novel approach that trains a joint ASR and SD system end-to-end, feeding the acoustic feature vector sequence into our RNN-T model, which outputs an interleaved sequence of words and speaker roles, as illustrated in Figure~\ref{fig:overview}. 

The two key insights we exploit in our work are: (a) RNN-Ts can output richer set of targets symbols such as speaker role and punctuation, since they allow prediction of symbols without the need to explicitly attach observations to them, and (b) they can seamlessly integrate acoustic and linguistic information. 

As a proof of concept, this article focuses on enriching the traditional speech recognition units with speaker roles, as illustrated in Figure~\ref{fig:overview}. The closest approach we could find in the literature is~\cite{Park2018}, which makes use of a sequence to sequence model, but only for SD and not ASR. For this task, we augment the output symbol set with new speaker role tokens. In the case of medical conversations between a physician and a patient, those additional tokens would e.g. be \texttt{<spk:dr>} and \texttt{<spk:pt>}. For a given audio input snippet consisting of a speech conversation between a physician and a patient, our RNN-T model predicts a joint ASR and SD output sequence, as illustrated by an example in Figure~\ref{fig:output-sequence}.

\subsection{Recurrent Neural Network Transducer}
\label{ssec:rnnt}
%TODO(soltau): rephrase ?
Let $\textbf{X} = \left( \textbf{x}_{1}, \textbf{x}_{2}, \ldots, \textbf{x}_{T}\right)$ be an input sequence of acoustic frames\footnote{As a naming convention, we employ uppercase bold symbols to denote variables that represent sequences over time, and corresponding lowercase bold symbols for elements within such a sequence at given time steps.}, where $T$ is the number of frames in the sequence. Typically, $\textbf{x}_{t} \in \mathbb{R}^{d}$ are log-mel filterbank energies. Let $\textbf{Y} = \left( \textbf{y}_{1}, \textbf{y}_{2}, \ldots, \textbf{y}_{U}\right)$ be the corresponding output sequence of symbols (including speaker roles) over the RNN-T output space $\mathcal{Y}$, and $\mathcal{Y}^{\ast}$ be the set of all possible sequence over $\mathcal{Y}$. To handle different alignments between input and output sequences, we define an extended output space $\mathcal{\bar{Y}} = \mathcal{Y} \cup \left\{\varnothing\right\}$, where $\varnothing$ denotes the blank output, as well as the corresponding set $\mathcal{\bar{Y}}^{\ast}$ of all possible sequences over $\mathcal{\bar{Y}}$. An element $\textbf{A} \in \bar{\mathcal{Y}}^{\ast}$ is referred to as an alignment, because the location of the blank symbols defines a mapping between the input and output symbols. For instance, the sequence $\textbf{A} = \left( \textbf{y}_{1}, \varnothing, \textbf{y}_{2}, \varnothing, \varnothing, \textbf{y}_{3}\right) \in \bar{\mathcal{Y}}^{\ast}$ is then equivalent to $\left( \textbf{y}_{1}, \textbf{y}_{2}, \textbf{y}_{3}\right) \in \mathcal{Y}^{\ast}$. Given the input sequence of frames $\textbf{X}$, the RNN-T model~\cite{Graves2012} defines a conditional probability $P\left(\textbf{Y} \in \mathcal{Y}^{\ast} \mid \textbf{X}\right)$ by marginalizing over the possible alignments

\begin{equation}
P\left(\textbf{Y} \in \mathcal{Y}^{\ast} \mid \textbf{X}\right) = \sum_{\textbf{A} \in \mathcal{B}^{-1}\left(\textbf{Y}\right)} P\left(\textbf{A} \mid \textbf{X}\right),
\end{equation}

\noindent where $\mathcal{B}$ is the function that removes the blank symbols from a given alignment in $\bar{\mathcal{Y}}^{\ast}$.

\begin{figure}
    \centering
    \input{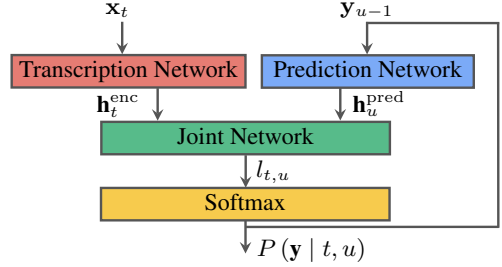}
    \caption{Architecture of the RNN-T-based model.}
    \label{fig:rnnt-architecture}
    \vspace{-0.5cm}
\end{figure}

An RNN-T models this conditional probability using three different networks. (1) A \textit{transcription network} (commonly called \textit{encoder} in the sequence to sequence literature) maps the acoustic frames $\left\{ \textbf{x}_{\tau} \right\}_{1 \leq \tau \leq t}$ into a higher level representation $\textbf{h}_{t}^{\mathrm{enc}} = f^{\mathrm{enc}}\left(\left\{ \textbf{x}_{\tau} \right\}_{1 \leq \tau \leq t}\right)$. (2) A \textit{prediction network} makes predictions based on the previous non-blank symbol such that $\textbf{h}_{u}^{\mathrm{pred}} = f^{\mathrm{pred}}\left( \left\{ \textbf{y}_{v} \right\}_{1 \leq v \leq u-1} \right)$. Finally, (3) a \textit{joint network} combines the output of the previous two networks to produce the logits $l_{t,u} = f^{\mathrm{joint}} \left(\textbf{h}_{t}^{\mathrm{enc}},\textbf{h}_{u}^{\mathrm{pred}}\right)$. The output of the joint network is then fed into a softmax layer to define a probability distribution. The model is optimized by maximizing the log-likelihood of $P\left(\textbf{Y} \in \mathcal{Y}^{\ast} \mid \textbf{X}\right)$.

%To keep the size of the output space $\bar{\mathcal{Y}}$ reasonable, we employ morphemes as elements of this set. The morphemes are then mapped back into a sequence of words $\textbf{W} = \left( \textbf{w}_{1}, \textbf{w}_{2}, \ldots, \textbf{w}_{V}\right)$ belonging to our vocabulary $\mathcal{W}$ using a look-up table. The resulting model architecture is summarized on Figure~\ref{fig:rnnt-architecture}.

The forward-back algorithm is used to calculate $P\left(\textbf{Y} \in \mathcal{Y}^{\ast} \mid \textbf{X}\right)$ efficiently via dynamic programming. Training the RNN-T model on accelerators like graphical processing units (GPU) or tensor processing units (TPU~\cite{Jouppi2017}) is non-trivial as computation of the loss function requires running the forward-backward algorithm. This issue was addressed recently in a TPU friendly implementation of the forward-backward algorithm, which recasts the problem as a sequence of matrix multiplications~\cite{Sim2017}. We took advantage of an efficient implementation of the RNN-T loss in TensorFlow that allowed quick iterations of model development~\cite{Bagby2018}. 

%While the naive calculation of $P\left(\textbf{Y} \in \mathcal{Y}^{\ast} \mid \textbf{X}\right)$ from all the possible alignments would be intractable, an efficient forward-back algorithm has been described in~\cite{Graves2012} and, later, variants were proposed to run efficiently on GPUs and TPUs~\cite{Sim2017}. Finally, the overall model is differentiable and can, hence, be trained end-to-end by maximizing the log-likelihood of the target using gradient descent.

%At inference time, a Gaussian mixture model-based voice activity detection splits the input signal input speech segments that are then fed into our RNN-T-based model. The decoding of the output sequence is then performed using a beam search for each segment separately.
% TODO(hagen): Extend this description a little bit.

\begin{figure}
    \centering
    \includegraphics[width=0.47\textwidth]{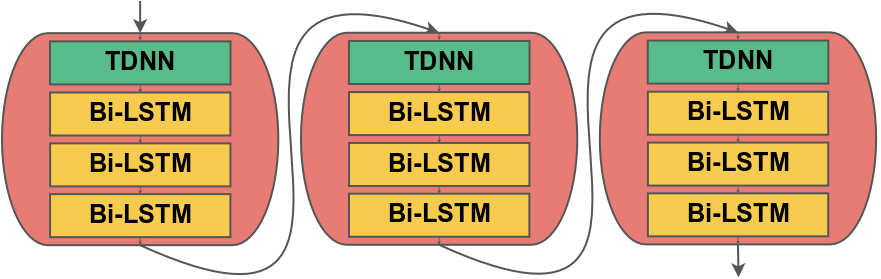}
    \caption{Transcription network (encoder) architecture}
    \label{fig:architecture-encoder}
    \vspace{-0.5cm}
\end{figure}

\subsection{Model Implementation}
\label{ssec:model-implementation}
%It is possible to rely on a wide variety of neural networks for the different RNN-T components. In practice, we leverage our past experience of training general purpose RNN-T-based ASR systems.

For training purposes, we split long conversations into audio segments of maximum $15$s that may contain multiple speakers. The corresponding output targets are speaker role decorated transcripts as previously depicted on Figure~\ref{fig:output-sequence}. We then extract acoustic frames, which are 80-dimensional logmel filterbank energies ($d=80$).

While sequence to sequence models often make use of graphemes as units, we argue that longer units are more appropriate for speech recognition. For example, if training data is abundant, entire words can be modeled directly in an LVCSR system~\cite{Soltau2017}. In this work, we choose a middle ground, and use morphemes as units that were obtained by a data driven approach~\cite{Virpioja2013}. Longer units have the advantage that we can reduce the time resolution for the output sequence, making both training and inference more efficient. We do this by employing a hierarchy of time delay neural network (TDNN) layers that reduces the time resolution from $10$ms to $80$ms~\cite{Waibel1995}. The architecture is very similar to the encoder used for CTC word models where the time resolution was reduced to 120ms and this decimation improved both inference speed and word error rate~\cite{Soltau2017b}.

Specifically, our encoder consists of three identical blocks made of four layers (see Figure~\ref{fig:architecture-encoder}): (a) a $1D$ temporal convolutional layer with $512$ filters, a kernel size of $5$ and a max-pooling operator of size $2$ followed by (b) three bi-directional long short-term memory (LSTM~\cite{Hochreiter1997}) layers with $512$ units. The prediction network consists of a word embedding layer that maps our morpheme vocabulary of about $4$K units to a $512$ dimension vector space followed by a uni-directional LSTM with $1024$ units and a fully connected layer with $512$ units.

%Our morpheme space consists of , while the size of our output vocabulary space $\mathcal{W}$ is $135,220$.

%TODO(soltau) : Add reference to Morfessor, add reference for TDNN
%TODO(soltau) : 8x time resolution via TDNN makes learning alignment easier, 
%TODO(soltau) : Architecture with interleaved lstm+tdnn similar to word model encoder, soltau2017
%We rely on a deep and high-capacity hybrid encoder, which consists of three identical blocks made of four layers (see Figure~\ref{fig:architecture-encoder}): (a) a $1D$ temporal convolutional layer with $512$ filters with a kernel size of $5$ and a max-pooling operator of size $2$~\cite{Waibel1995} followed by (b) three bi-directional long short-term memory (LSTM)~\cite{Hochreiter1997} layers with $512$ units. The prediction network consists of a word embedding layer, a uni-directional LSTM with $512$ units and a fully connected layer with $512$ units. Finally, the joint network is made of a single fully connected layer with as many units as elements in our output symbol space $\mathcal{Y}^{\ast}$. The size of our morpheme output space $\mathcal{Y}^{\ast}$ inclusive special tokens such as the speaker labels is $4,135$, while the size of our output vocabulary space $\mathcal{W}$ is $135,220$.
%TODO(soltau) : this is training vocab, for inference we have open vocab

Training is performed using the stochastic gradient-based Adam optimizer~\cite{Kingma2014}. We train the model on a set of $128$ TPUs and it converges in less than two days.

\section{Experiments}
\label{sec:expts}

\subsection{Corpus}
\label{ssec:corpus}
We experimented on a large corpus of about $100$K ($\approx 15$K hours) manually transcribed audio recordings of clinical conversations between physicians and patients, where each conversation is about $10$ minutes long on the average. The transcription breaks up a conversation into speaker turns and in each turn identifies the speaker role label and the sequence of words spoken. This long-form transcription makes the corpus well-suited for training our proposed model. Each conversation typically contains two different speaker roles, respectively \texttt{<spk:dr>} and \texttt{<spk:pt>}. For the relatively few cases where other speakers are involved, we map them to the closest speaker role (e.g., nurses to \texttt{<spk:dr>}, family caregivers to \texttt{<spk:pt>}). The acoustic quality of the recordings varied significantly due to several factors -- the distance of the speakers to the microphone, recording devices, the audio encoder (mp3 or wav) and the sample rate ($8$kHz or $16$kHz). 
%The conversations are de-identified in the sense that protected health information (PHI) data such as names and dates of birth is wiped out of the conversations by (1) zeroing the corresponding audio segment and (2) removing the corresponding words from the transcript. 
We partitioned our corpus into three sets, holding out $508$ and $404$ conversations for development and evaluation sets respectively, and the rest for training. There are no overlaps of physicians between the three sets. The patient overlap is uncertain since their identities are unavailable. The training data is split into segments of less than $15$sec, as mentioned earlier, and results in an average of about $4$ speaker turns per segment.
%TODO(shafey): retrieve the exact number of training segments

\subsection{Evaluation Metric}
\label{ssec:metrics}
The common approach to evaluate conventional diarization systems is to rely on the diarization error rate ($\mathrm{DER}$) metric, which compares reference speaker-labeled segments with SD predictions in the time domain. In contrast, the use of a joint ASR and SD system directly assigns speaker roles to the recognized words, and hence makes it unnecessary to depend on the time boundaries for aligning words with the speaker roles.

Motivated by the speaker attributed task proposed in the NIST RT-03F evaluation plan~\cite{RT2003} (Section~5.2.5), we utilize a metric suitable to assess such end-to-end joint ASR and SD systems, by measuring the percentage of words in the transcript decorated with the right speaker tag. Specifically, we define the Word Diarization Error Rate ($\mathrm{WDER}$) as:

\vspace{-0.25cm}
\begin{equation}
    \mathrm{WDER} = \frac{S_{\mathrm{IS}} + C_{\mathrm{IS}}}{S + C}
\end{equation}
\vspace{-0.2cm}
\noindent where,
\begin{enumerate}
    \setlength{\itemsep}{0pt}
    \setlength{\parskip}{0pt}
    \item $S_{\mathrm{IS}}$ is the number of ASR Substitutions with Incorrect Speaker tokens,
    \item $C_{\mathrm{IS}}$ is the number of Correct ASR words with Incorrect Speaker tokens,
    \item $S$ is the number of ASR substitutions,
    \item $C$ is the number of Correct ASR words.
\end{enumerate}
\noindent Note that this $\mathrm{WDER}$ metric must be used in combination with the ASR Word Error Rate ($\mathrm{WER}$) to account for deletions and insertions since the speaker labels associated with them cannot be mapped to reference without ambiguity.
In our opinion, this word-level metric reflects the performance in an actual application better than the time-level metric.

\subsection{Baseline}
\label{ssec:baseline}

Since our proposed joint model is based on RNN-T, we compare its performance with a baseline system that also uses an equivalent RNN-T model but only for ASR. The architecture of the ASR system is same as described in Section~\ref{ssec:model-implementation}, except that the speaker roles are removed from the transcript. Based on our past experience with conventional SD system, we built a strong baseline system consisting of the following five stages:

% TODO(izhak): Need to add citations
\noindent (a)~\textbf{Speech detection and segmentation}: This stage consists of an LSTM-based speech detector whose threshold is kept low to minimize deletion of speech segments~\cite{Zazo2016}.

\noindent (b)~\textbf{Speaker embedding}: The speaker embeddings are computed using a sliding window of $1$sec with a stride of $100$ms. An acoustic feature sequence is extracted from the $1$sec window and fed into an LSTM model. The last hidden state of the LSTM is extracted as the embedding for the $1$sec window ~\cite{Heigold2016}. The embeddings are trained to maximize classification of speakers in the training set~\cite{Snyder2018,Sell2018}. 
%and on the union of the two. In practice, we obtained the best performance on the development set using VoxCeleb2 data only. We believe this is caused by (1) the higher quality of the VoxCeleb2 data (e.g. higher sample rate) and (2) the higher accuracy of the speaker segment boundaries in the time domain.

\noindent (c)~\textbf{Speaker change detection}: Cosine distance is computed between adjacent embedding vectors in the speech segments. When the distance is higher than a threshold (optimized on the development set), the transition is marked as a speaker change. Thus, speech segments are broken into single speaker segments.

\noindent (d)~\textbf{Speaker clustering}: The single speaker segments are clustered so that the speaker labels can be applied consistently across all segments from the same speaker. Among various choices of clustering algorithms, we found k-means to be most effective, with $k=2$.

\noindent (e)~\textbf{Reconciling the ASR and SD output}: The SD system provides speaker turns with time boundaries and these labels are mapped to recognized words using the associated word boundaries from the ASR system. When the speaker turn boundary fall in the middle of a word, we assign the word to the speaker with the largest overlap with the word. The baseline predicts generic speaker tags such as \texttt{<spk:0>} and \texttt{<spk:1>}.

For evaluation purposes, we map the generic labels back to speaker roles on a per conversation basis. The mappings are picked to minimize the diarization errors, hence, giving an advantage to the baseline over the proposed system. For training speaker embeddings, we augmented our training data with the VoxCeleb2 dataset~\cite{Chung2018} since it improved the performance over using only our corpus. Note, the VoxCeleb2 data does not contain long-form transcription or speaker role labels in medical domain and hence cannot be used for training the integrated RNN-T models. Thus, the baseline used more data than our RNN-T model.

\subsection{Results and Analysis}
\label{ssec:results}

In Table~\ref{tab:results-wder}, we compare the performance of our system with the strong baseline described above. We observe a substantial improvement in $\mathrm{WDER}$, which drops from $15.8\%$ to $2.2\%$, a relative improvement of about $86\%$ over the baseline. This gain in $\mathrm{WDER}$ comes at a small cost in ASR performance with about $0.6\%$ degradation in $\mathrm{WER}$.

\begin{table}[b]
   \small % text size of table content
   \centering % center the table
   \begin{tabular}{lcc} % alignment of each column data
   \toprule[\heavyrulewidth]\toprule[\heavyrulewidth]
    & \textbf{Baseline} & \textbf{Joint ASR+SD} \\ 
   \midrule
   $\mathrm{WDER}$ & $15.8\%$ & $\mathbf{2.2\%}$ \\
   \hdashline
   $\mathrm{WER}$ &  $\mathbf{18.7\%}$ & $19.3\%$ \\
   $\mathrm{D} / \mathrm{I} / \mathrm{S}$ &  $7.2\% / 2.1\% / 9.4\%$ & $6.8\% / 2.8\% / 9.7\%$ \\
   \bottomrule[\heavyrulewidth] 
   \end{tabular}
   \caption{Word Diarization Error Rate ($\mathrm{WDER}$), Word Error Rate ($\mathrm{WER}$) and its decomposition in Deletion / Insertion / Substitution errors ($\mathrm{D} / \mathrm{I} / \mathrm{S}$) on the evaluation set.} 
   \label{tab:results-wder}
\end{table}

%This implies that our augmented ASR-like model is able to accurately predict speaker roles, which, as opposed to words, correspond to transitions rather than actual phonemes.

\begin{figure}
    \centering
    \includegraphics[width=0.45\textwidth]{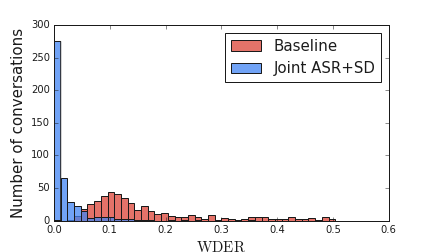}
    \caption{Distribution of the $\mathrm{WDER}$ on a per conversation basis for the baseline and the proposed system.}
    \label{fig:wder-dist}
    \vspace{-0.5cm}
\end{figure}

In a conventional system, the $\mathrm{WDER}$ is affected by several factors such as errors in ASR generated word boundaries, errors in SD generated speaker turn boundaries as well as errors in the ad hoc reconciliation step. In contrast, our proposed system clearly benefits from the lack of such intermediate steps.
%The $\mathrm{WER}$ of about $19\%$ may look high. But our corpus consists of natural conversations between physicians and patients with a lot of mumbling, unintelligible or overlapping speech. It is, hence, expected to have a higher $\mathrm{WER}$ than in e.g. broadcast news or television interviews.

Examining the distribution of errors across conversations, as shown in Figure~\ref{fig:wder-dist}, we find that the performance of our proposed system is much more consistent with most of the conversations under $\mathrm{WDER}$ of $4\%$, while the baseline system has a wide spread around a mode of $11\%$. On manual inspection of the outliers of our system (above $15\%$), interestingly, the speaker change detections rarely fail, but, once the model wrongly assigns speaker roles, they stay flipped for a large segment of the conversation. We suspect this is due to the way inference is performed on chunks of conversation in batches and not on the entire conversation.

\begin{figure}
    \centering
    \includegraphics[width=0.45\textwidth]{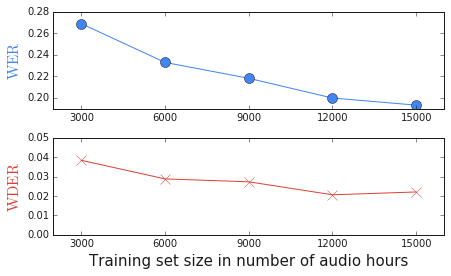}
    \caption{Impact of amount of training data on the performance of the proposed system on the evaluation set.}
    \label{fig:impact-training-data}
    \vspace{-0.5cm}
\end{figure}

Joint models are notorious for being susceptible to data sparsity, so we investigate the impact of training data size on our model performance. In the Figure~\ref{fig:impact-training-data}, we plot the performance with respect to training data size ranging from $3,000$ to $15,000$ hours of audio. For ease of experimentation, we didn't tune the model size. We observe that more training data seems to be more helpful to improve $\mathrm{WER}$ rather than $\mathrm{WDER}$.

\section{Conclusions And Future Work}
\label{sec:conclusions}

We introduced a novel joint ASR and SD system, which relies on the sequence to sequence paradigm and is implemented using an RNN-T model. We demonstrated the performance of our approach by evaluating it on a large corpus of clinical conversations between physicians and patients. Compared to a conventional baseline, we observed a significant relative improvement of $86\%$ in the word-level diarization error rate, without significant degradation in the word error rate.

This system is particularly well-suited for applications where there are limited number of speakers in a conversation and speaker roles fall into well-defined categories. Note that the focus of our system is labeling speaker roles rather than identities, and as such it is critical to have matched speaker role labeled training data. Unfortunately, to the best of our knowledge, there is no public corpus where we could have evaluated our model for the benefit of the larger community. One other question that we were unable to address in this work is how much of the performance gain was from the lexical cues. This is complicated by the fact that lexical cues need to be inferred on spoken words and not transcribed ones.

In the future, we would like to evaluate our approach on other applications with clear speaker roles. Longer term, we are interested in providing truly rich transcripts of conversations. For example, we are currently conducting experiments to include punctuation and capitalization that look promising and we believe that the approach could be highly beneficial for exploiting non-verbal cues such as emotions.

%of such a technique when there are more than two speaker roles as well as on other types of scenarios such as broadcast news or television interviews.
%Longer term, we are interested in a generalization of this approach to the more generic speaker diarization problem, and in determining if there is a benefit to tackle few of the natural language understanding problems from the raw audio rather than the actual automatic transcript.

\section{Acknowledgements}
We are grateful to Rick Rose and Olivier Siohan for many discussions and help with the baseline system, the $\mathrm{WDER}$ metric and its implementation, and to Gang Li for help with improving the speaker embedding for the baseline system.

\newpage
\bibliographystyle{IEEEtran}
\bibliography{rnnt_diarization}

\end{document}